\documentclass{article}

\usepackage[final, nonatbib]{neurips_2024_ml4ps}

\usepackage[utf8]{inputenc} 
\usepackage[T1]{fontenc}    
\usepackage{hyperref}       
\usepackage{url}            
\usepackage{booktabs}       
\usepackage{amsfonts}       
\usepackage{nicefrac}       
\usepackage{microtype}      
\usepackage{xcolor}         
\usepackage{graphicx}
\usepackage{amsmath}
\usepackage{subcaption}
\usepackage[numbers]{natbib}

\title{Harnessing Loss Decomposition for Long-Horizon Wave Predictions via Deep Neural Networks}

\author{%
  Indu Kant Deo \\ 
  Department of Mechanical Engineering\\ 
  The University of British Columbia\\
  Vancouver, BC V6T 1Z4\\
  \texttt{indukant@mail.ubc.ca} \\
  \And
    Rajeev K. Jaiman \\
Department of Mechanical Engineering\\ 
The University of British Columbia\\
Vancouver, BC V6T 1Z4\\
  \texttt{rjaiman@mail.ubc.ca} \\
}

\begin{document}

\maketitle

\begin{abstract}
Accurate prediction over long time horizons is crucial for modeling complex physical processes such as wave propagation. Although deep neural networks show promise for real-time forecasting, they often struggle with accumulating phase and amplitude errors as predictions extend over a long period. To address this issue, we propose a novel loss decomposition strategy that breaks down the loss into separate phase and amplitude components. This technique improves the long-term prediction accuracy of neural networks in wave propagation tasks by explicitly accounting for numerical errors, improving stability, and reducing error accumulation over extended forecasts.
\end{abstract}

\section{ Introduction}
Accurate long-horizon predictions are essential for understanding and modeling complex physical phenomena, particularly in the context of wave propagation governed by partial differential equations (PDEs) \cite{leveque2002finite}. 
 Traditional numerical methods for solving PDEs, such as finite difference, finite element, and spectral methods, have been extensively studied and applied \cite{anderson1995computational} to solve the wave equation. 
Although these methods are effective, they often require substantial computational resources, making them unsuitable for real-time predictions and control related tasks.

Machine learning techniques, particularly deep neural networks, have emerged as powerful tools for modeling complex physical phenomena, including wave propagation and underwater noise prediction \cite{deo2022predicting, deo2023combined, gao2024finite, deo2024predicting, deo2024continual}. 
Despite their success in numerous domains, neural networks often struggle with accurate long-term predictions, primarily due to accumulated phase and amplitude errors \cite{LeCun2015,Goodfellow2016,karpatne2017theory}. This issue is particularly pronounced in autoregressive models, where errors can compound over time, leading to significant deviations from the true dynamics \cite{Rottmann2018,beucler2021enforcing}.

Recent studies have highlighted the importance of loss decomposition in addressing these challenges. Traditional mean squared error (MSE) loss functions do not distinguish between different types of prediction errors, such as phase and amplitude errors, which are critical to accurate long-term forecasting \cite{le2019shape, wen2023reduced,Bai2018}. To mitigate this, we propose decomposing the MSE into dissipation and dispersion errors \cite{Griffiths_2016}, which separately account for the amplitude and phase errors, respectively. By explicitly considering these error types, our approach aims to enhance the neural network's ability to maintain accurate predictions over extended horizons.

The concept of loss decomposition is not entirely new; it has been previously explored in numerical analysis and signal processing. For example, Kreiss and Oliger \cite{Kreiss1972} introduced the idea of dissipation and dispersion errors in the context of finite difference methods. More recently, Guen et al. \cite{le2019shape} applied similar ideas to machine learning, suggesting that decomposed loss functions can guide neural networks to learn more robust representations of time series forecasting. Our work builds on these ideas by specifically targeting the phase and amplitude errors that hinder long-term predictions in neural networks for wave propagation tasks.
By integrating the perspective of numerical errors into machine learning techniques, we introduce a novel theoretical framework aimed at improving long-term predictions.

\section{Methods}
\subsection{Linear advection equation}
\begin{figure}[ht]
    \centering
    \includegraphics[width=0.99\linewidth]{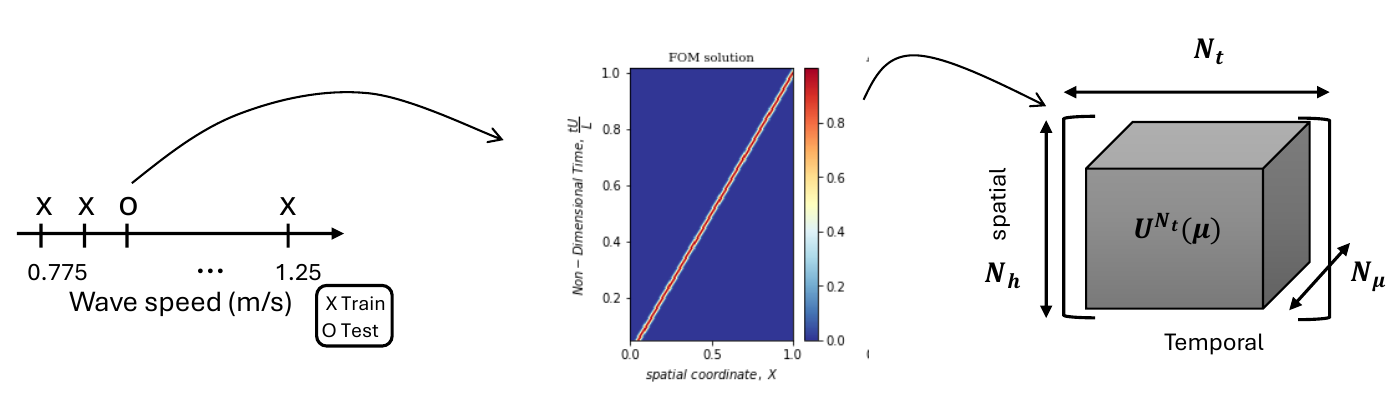}
    \caption{Schematic of dataset generation. Linear convection simulations were generated for parameter wave speed. Amplitude data were extracted and arranged into a training snapshot matrix.}
\label{fig:fig1}
\end{figure}

In this study, we focus on data generated from the linear convection equation as a model problem for wave propagation. Although the approach has been applied to the inviscid Burgers equation to investigate nonlinear wave propagation, we limit our discussion to the linear convection equation for brevity. The solution $U$ in the domain $\Omega \equiv [0, 1]$ satisfies the following parametric partial differential equation:
\begin{equation}
\frac{\partial U}{\partial t} + \mu \frac{\partial U}{\partial X} = 0 \quad \text{in~} \quad \Omega, 
\end{equation}  
with the initial condition: 
\begin{align} U(x, 0) = U_0(x) \equiv f(x), 
\end{align} where $\mu \in [0.775, 1.25]$ represents the wave speed. The initial profile is given by $f(x) = \left( \frac{1}{\sqrt{2\pi\sigma}} \right) e^{-x^2 / 2\sigma}$ with $\sigma = 5\times10^{-3}$. A one-dimensional spatial discretization of 256 grid points and 200 time steps is applied. The training set consists of $N_{\mu} = 20$ parameter instances, uniformly distributed over the range of $\mu$, while the test set includes $N_{\text{test}} = 19$ parameter instances. The test parameters are selected such that $\mu_{\text{test}, i} = \frac{\mu_{\text{train}, i} + \mu_{\text{train}, i+1}}{2}$ for $i = 1, \ldots, N_{\text{test}}$.
A visual representation of this dataset is provided in Figure \ref{fig:fig1}.

\subsection{Attention-based Convolution Recurrent Autoencoder Network}
\begin{figure*}
    \centering
    \includegraphics[width=\textwidth]{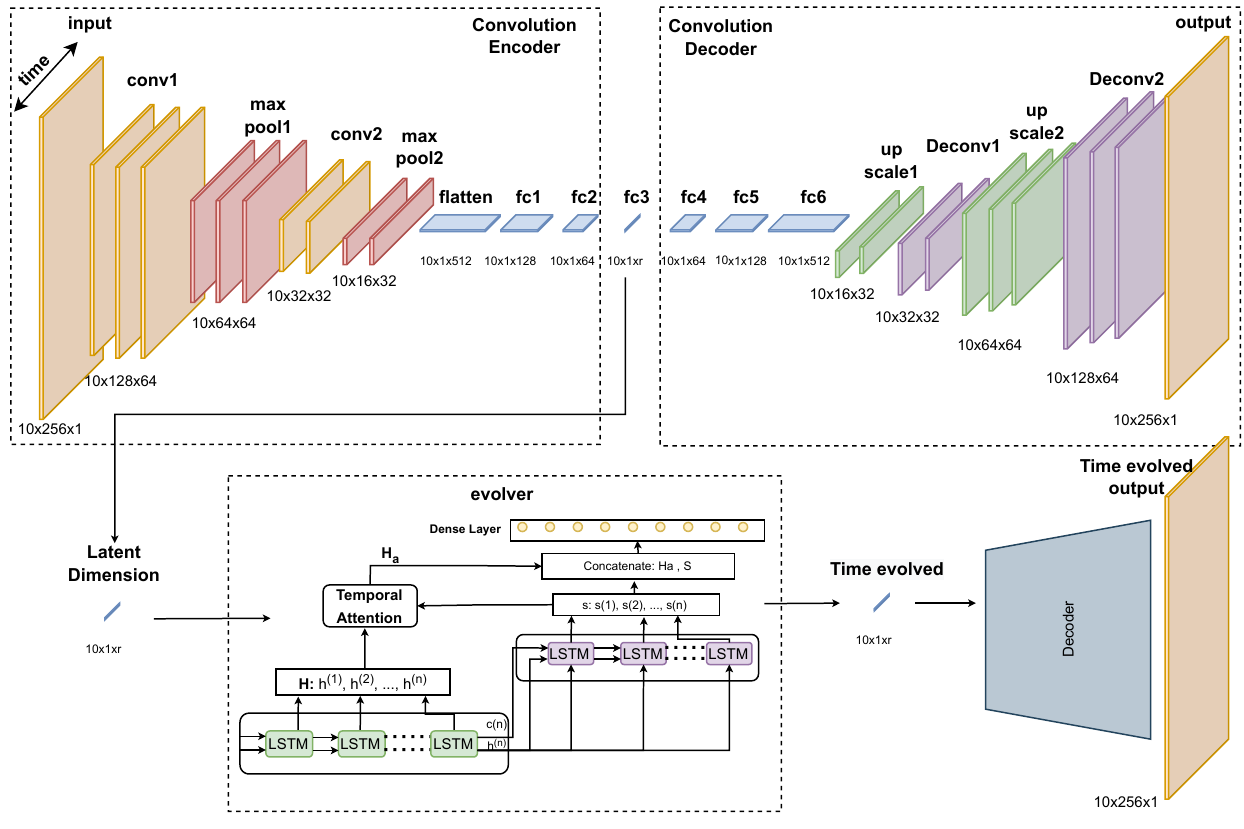}
    \caption{Illustration of attention-based convolutional recurrent autoencoder architecture. }
\label{fig:attention_CRAN_Lin}
\end{figure*}

We employ a deep learning architecture to model wave propagation, specifically utilizing a 15-layer Attention-Based Convolutional Recurrent Autoencoder Network \cite{deo2022predicting, gonzalez2018deep}. The encoder includes a convolutional layer, followed by max pooling and fully connected layers. The output layer of the encoder contains {$r$} neurons, where {$r$} represents the dimensionality of the reduced manifold. 
The propagator is composed of attention-based sequence-to-sequence RNN-LSTM layers that process the reduced manifold data, advancing it along the time dimension. The decoder, designed as the reverse of the encoder, reconstructs the data from the latent space back to the original physical dimension. The model architecture and hyperparameters were adopted from the work of Deo et al. \cite{deo2022predicting}.  The model is trained from scratch using TensorFlow \cite{tensorflow2015-whitepaper} on a single NVIDIA P2200 Pascal GPU, with training converging in 285 epochs within 20 minutes of wall-clock time.
Figure \ref{fig:attention_CRAN_Lin} shows the visual representation of AB-CRAN architecture.

\subsection{Loss decomposition}
Mean squared error (MSE) is commonly employed as the default loss function for regression tasks. However, MSE fails to account for numerical dissipation and dispersion errors in the solution. To address this limitation, MSE can be decomposed into two components: one representing dissipation error (amplitude) and the other capturing dispersion error (phase). The total mean square error can be rewritten as:
\begin{equation}
\tau=\sigma^2\left(u_a\right)+\sigma^2\left(u_d\right)-2 \rho \sigma\left(u_a\right) \sigma\left(u_d\right)+\left(\bar{u}_a-\bar{u}_d\right)^2,
\label{eqn:mse-bias-variance}
\end{equation}
where $u_a$ is the ground-truth solution and $u_d$ the predicted solution. $\sigma^2\left(u_a\right),  \sigma^2\left(u_d\right)$ are the variances of $\mathrm{u}_{\mathrm{a}}$ and $\mathrm{u}_{\mathrm{d}}$, respectively. $\operatorname{cov}\left(u_a, u_d\right)$ is the covariance between $u_a$ and $u_d$ and $\rho$ is the correlation coefficient.
Eq. \ref{eqn:mse-bias-variance} can be rewritten as

\begin{equation}
\tau=\left[\sigma\left(u_a\right)-\sigma\left(u_d\right)\right]^2+\left(\bar{u}_a-\bar{u}_d\right)^2+2(1-\rho) \sigma\left(u_a\right) \sigma\left(u_d\right).
\label{eqn:mse-BV}
\end{equation}

We define the first two terms of the RHS of Eq. \ref{eqn:mse-BV} as the dissipation error and the third term as the dispersion error:
\begin{equation}
\begin{aligned}
 \tau_{\text {DISS }} &= \left[\sigma\left(u_a\right)-\sigma\left(u_d\right)\right]^2+\left(\bar{u}_a-\bar{u}_d\right)^2,\\ 
\tau_{\text {DISP }} &= 2(1-\rho) \sigma\left(u_a\right) \sigma\left(u_d\right).
\label{eq:mse-final-b}
\end{aligned}
\end{equation}

When two wave patterns differ solely in amplitude but share the same phase, their correlation coefficient, $\rho$, should be 1, with $\tau_{\text{DISP}}=0$ as indicated by equation (\ref{eq:mse-final-b}) . This outcome aligns with our expectations and is a reasonable result. We employ this decomposition to formulate the loss function as follows:
\begin{equation}
\begin{aligned}
 &\mathcal{L}_{\text {function }} = (1-\alpha)\mathcal{L}_{\text {Decoder }} + \alpha \mathcal{L}_{\text {propagator }},\\ 
&\mathcal{L}_{\text {propagator }} =  (1-\beta)\tau_{\text {DISP }} + \beta\tau_{\text {DISS }},
\label{eq:loss}
\end{aligned}
\end{equation}
where $\alpha$ and $\beta$ are hyperparameters. To determine the values of $\alpha$ and $\beta$, we utilize hyperparameter tuning with the Ray Tune ASHA algorithm \cite{li1810system}. Our experiments reveal that setting $\alpha$ and $\beta$ greater than 0.5 prioritizing the propagator, with a stronger emphasis on reducing dispersion error within the propagator works best.
This work presents the first introduction of phase and amplitude components into the loss function for training deep neural networks for wave propagation.
\subsection{Training}

To improve the training of the attention-based convolutional recurrent autoencoder (AB-CRAN), we integrate a denoising mechanism into the decoder's loss function. During the initial forward pass, noise is added to the input data before passing it through the encoder-decoder pipeline. The decoder is then tasked with reconstructing the original noise-free data, optimizing the mean squared error between the noisy inputs and the clean data, resulting in a denoising decoder loss. In the second stage, the projected data from the encoder are processed through the propagator to generate a time-evolved output, and the corresponding loss is decomposed into dissipation and dispersion errors, as explained in the above section. The entire model is optimized using the AdamW optimizer \cite{zhuang2022understanding} and a cosine annealing warm-restart scheduler \cite{loshchilov2016sgdr}. We also implemented an early stopping strategy with a patience threshold of 50 epochs.
\section{Results}
\begin{figure*}
\centering
\includegraphics[width=0.99\textwidth]{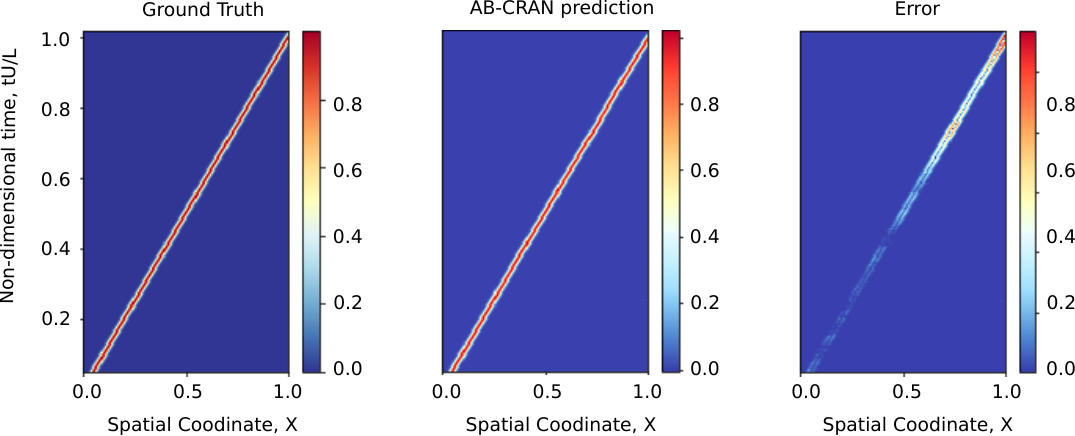}
 \caption{Linear convection problem: Exact solution (left), AB-CRAN solution with n = 2 (center) and error $ e = |\hat{u} - u|$ (right) for the testing parameter $\mu_{test}$ = 1.0125 in the space-time domain.}
\label{fig:char_ABCRAN_mu_0}
\end{figure*}

We randomly selected a parameter from the test set for evaluation. The AB-CRAN architecture was provided with data from the linear convection equation, using $\mu = 1.0125$ as input. Figure~\ref{fig:char_ABCRAN_mu_0} illustrates the exact solution alongside the AB-CRAN approximation for this test case. The results demonstrate that the AB-CRAN model, trained with the proposed loss decomposition strategy, captures wave dynamics with high fidelity, accurately preserving key features of the solution.

\begin{figure}[htbp]
    \centering
    \begin{subfigure}{0.8\textwidth}
        \centering
        \includegraphics[width=\textwidth]{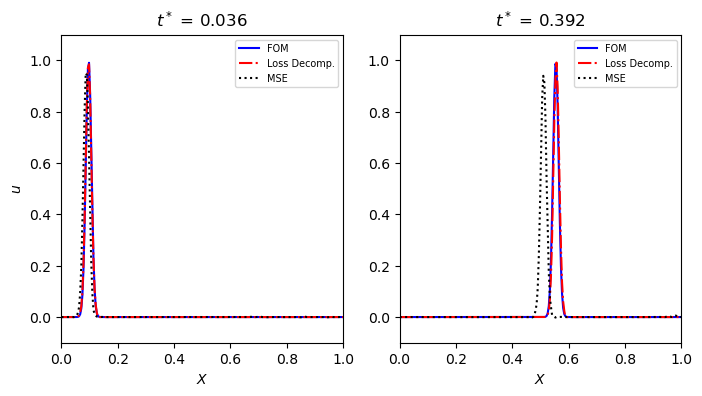}
        \caption{}
        \label{fig:subfig1}
    \end{subfigure}
    \hfill
    \begin{subfigure}{0.8\textwidth}
        \centering
        \includegraphics[width=\textwidth]{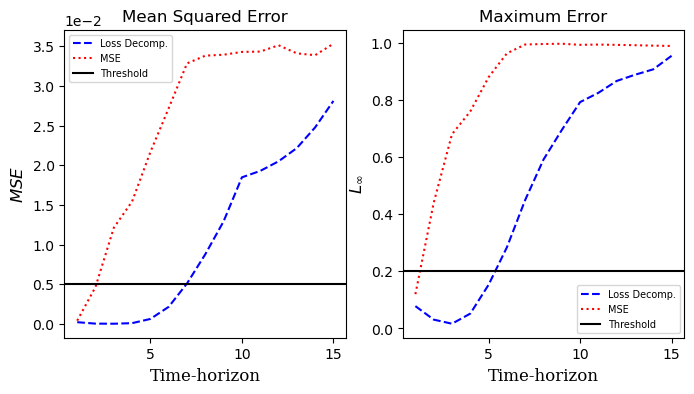}
        \caption{}
        \label{fig:subfig2}
    \end{subfigure}
    \caption{Accuracy assessment for wave prediction: (a) Comparison of MSE-based loss and Loss-decomposition-based training prediction for $t^* $= 0.036,  0.392  (b) Error vs. time-horizon plot comparing MSE-based and loss-decomposition-based training predictions, with a time-horizon of 10 time-steps}
    \label{fig:side_by_side}
\end{figure}

Our loss decomposition strategy significantly improves phase accuracy compared to the conventional mean squared error (MSE) loss. As shown in Figure~\ref{fig:subfig1}, predictions based on MSE exhibit noticeable phase lag over time, highlighting the inability of traditional loss functions to maintain phase alignment. In contrast, our method preserves phase throughout the prediction horizon. Furthermore, Figure~\ref{fig:subfig2} demonstrates that the proposed approach extends the prediction horizon by effectively mitigating error accumulation. By separately addressing phase and amplitude errors, the model achieves greater stability and accuracy, particularly for long-term forecasts.

This decomposition-based approach provides a robust and generalizable framework for long-term forecasting of physical systems. By isolating and addressing phase and amplitude errors, it overcomes limitations inherent in traditional MSE-based methods. The strategy reduces error accumulation, ensuring phase alignment and amplitude stability over extended time horizons. These improvements make the method broadly applicable to various physical forecasting tasks, such as wave propagation, fluid dynamics, and climate modeling, establishing a new standard for accuracy and reliability in long-horizon predictions.

\section{Conclusion}
In this work, we presented a novel loss decomposition strategy that addresses the limitations of traditional mean squared error (MSE)-based loss functions for long-term predictions in physical systems. By separating the loss into phase and amplitude components, our approach independently tackles dispersion errors (phase misalignment) and dissipation errors (amplitude inaccuracies). This targeted decomposition significantly enhances both the accuracy and stability of predictions, extending the prediction horizon while maintaining physical consistency. Our results demonstrate that the proposed method outperforms standard MSE-based approaches, offering substantial improvements for tasks such as wave propagation modeling.

The flexibility of this framework makes it broadly applicable to various forecasting tasks in physical domains, including flow dynamics, ocean acoustics, and climate modeling. By addressing phase and amplitude errors independently, our method ensures reliable long-term predictions for complex systems. This study underscores the potential of tailored loss functions to advance the modeling and simulation of complex dynamical systems, providing a promising foundation for future research in data-driven physical modeling and scientific computing.

\bibliographystyle{plainnat}
\bibliography{references}

\end{document}